\documentclass[letterpaper, 10 pt, conference]{ieeeconf}  

\IEEEoverridecommandlockouts                              
\title{\LARGE \bf
Learning Classifiers for Imbalanced and Overlapping Data
}
\usepackage[utf8]{inputenc}
\usepackage{graphicx}
\usepackage{authblk}
\author[1]{Pranav Murali${ 19BPS1035}$}
\author[2]{Nitin Narayanan N${ 19BPS1050}$}
\author[3]{Ajaykumar M${ 19BPS1093}$}
\author[4]{Shivaditya Shivganesh${ 19BPS1103}$}
\author{Shivaditya Shivganesh\\19BPS1103 \and Nitin Narayanan N\\19BPS1050 \and Pranav Murali\\19BPS1035 \and Ajaykumar M\\19BPS1093}

\begin{document}

\maketitle
\thispagestyle{empty}
\pagestyle{empty}

\begin{abstract}

This study is about inducing classifiers using data that is imbalanced, with a minority class being under-represented in relation to the majority classes. The first section of this research focuses on the main characteristics of data that generate this problem. Following a study of previous, relevant research, a variety of artificial, imbalanced data sets influenced by important elements were created. These data sets were used to create decision trees and rule-based classifiers. The second section of this research looks into how to improve classifiers by pre-processing data with resampling approaches. The results of the following trials are compared to the performance of distinct pre-processing re-sampling methods: two variants of random over-sampling and focused under-sampling NCR. This paper further optimises class imbalance with a new method called Sparsity. The data is made more sparse from its class centers, hence making it more homogenous.
\end{abstract}

\section{INTRODUCTION}

Supervised learning of classifiers from examples is one of the main tasks in machine learning and data mining.  However, their usefulness for obtaining high predictive accuracy in real life data depends on different factors, including also difficulties of the learning problem and its data characteristics. Class imbalance is one of the sources of these difficulties. Many real life problems are characterized by a highly imbalanced distribution of examples in classes. Typical examples are rare medical diagnosis, recognition of oil spills in satellite images, detecting specific astronomical objects in sky surveys or technical diagnostics of equipment failures. Moreover, in fraud detection, either in card transactions or in telephone calls  the number of legitimate transactions is much higher than the number of fraudulent ones. Similar situations occur either in direct marketing where the response rate class is usually very small in most marketing campaigns or information filtering where some important categories contain few messages only.

\subsection{Issues with Class Imbalance}
If imbalance in the class distribution is extensive, i.e. some classes are strongly under-represented, then the typical learning methods do not work properly. An even class distribution is often assumed (also non explicitly) and the classifiers are “somehow biased” to focus searching on the more frequent classes while “missing” examples from the minority class. As a result constructed classifiers are also biased toward recognition of the majority classes and they usually have difficulties (or even are unable) to classify correctly new objects from the minority class.

\subsection{Possible Solutions}
This paper concerns classifier-independent methods that rely on transforming the original data to change the distribution of classes, e.g., by resampling as these methods are more universal and they can be used in a pre-processing stage before applying many learning algorithms. A small number of examples in the minority class is not the only source of difficulties for classifiers. Recent works also suggest that there are other factors that contribute to difficulties.  The degradation of performance was also related to other factors, mainly to decomposition of the minority class into many sub-clusters with very few examples. The rare sub-concepts correspond to, so called, small disjuncts, which lead to classification errors more often than examples from larger parts of the class.

\subsection{Goals}
Studying the role of these factors in class imbalance is still an open research problem. Therefore, the main aim of this study is to experimentally examine which of these factors are more critical for the performance of the classifier. Carrying out such experiments requires preparing a new collection of artificial data sets which are affected by the above mentioned factors. Proposing such data sets is another sub-aim of this paper. In this paper we are particularly interested in focused (also called informed) resampling methods, which modify the class distribution taking into account local characteristics of examples. Representative methods such as SMOTE for selective over-sampling of the minority class, one side sampling and NCR for removing examples from the majority classes are used.

\section{Literature Review}

Jerzy Stefanowski works on inducing classifiers from unbalanced data, in which one class (a minority class) is underrepresented compared to the other classes (majority classes). The minority group is usually the focus of attention, and it must be recognised as accurately as possible. Because most algorithms learning classifiers are biased toward the majority classes, class imbalance is a challenge for them. The first section of this research focuses on the main characteristics of data that generate this problem. Following the review, related research resulted in the creation of numerous forms of artificial, imbalanced data sets influenced by important aspects. These data sets were used to create decision trees and rule-based classifiers. The results of the initial studies suggest that a lack of examples from the minority class is not the primary cause of problems.These findings support the initial premise that the loss of classification performance is mainly related to the disintegration of minority classes into small sub-parts, and we base our work on these findings, employing a variety of strategies to address these concerns.

Chawla, N. V., et al offer a method for constructing classifiers from unbalanced datasets. If the categorization categories are not roughly equally represented, the dataset is unbalanced. "Normal" examples make up the majority of real-world datasets, with only a small number of "abnormal" or "interesting" examples. It's also true that the cost of misclassifying an unusual (interesting) sample as a normal example is frequently significantly higher than the cost of making the opposite mistake. Under-sampling the majority (normal) class has been offered as a good way to improve a classifier's sensitivity to the minority class; this strategy has also been considered and tried alongside other strategies.

\section{Analysis}

If one of the target classes has a significantly lower number of examples in classes than the other, the dataset is called imbalanced. Minority courses are those who are underrepresented, while majority classes are those who are over-represented. The table below shows the results. When a sub-cluster dataset is subjected to an imbalance ratio of (0.2), the model performance worsens as the number of minority class sub clusters grows. Two of the primary elements for this degradation are shown in the table below. One of them is sample size; as the sample size shrinks, so does the model's performance. As a result, both the imbalance ratio and sample size play a role in creating an unbalanced sample.

\subsection{Model Performance}

\begin{tabular}{||c c c c||}
 \hline
 {Number of Sub Clusters} & \multicolumn{3}{|c|}{1:5} \\
  & 600 & 400 & 200 \\ [0.5ex]
 \hline\hline
 2 & 0.803 & 0.845 & 0.742 \\
 \hline
 3 & 0.767 & 0.765 & 0.712\\
 \hline
 4 & 0.787 & 0.712 & 0.721\\
 \hline
 5 & 0.742 & 0.694 & 0.655\\
 \hline
 6 & 0.651 & 0.684 & 0.59\\ [1ex]
 \hline
\end{tabular}

\subsection{Generating Imbalanced Dataset}

In order to solve to the issue of class imbalance in real world problems we need to test and emulate the situations in order to devise solutions. So, we have used an isotropic gaussian distribution in order to randomly sample data points from it. The randomly sampled points are used to plot the cluster centers with which we can make cluster blobs. Apart from the isotropic distribution binomial or exponential distribution can also be used for random sampling. We generate cluster centers based on which we create data points in proximity to the cluster center in the coordinate space.

\begin{figure}[htp]
    \centering
    \includegraphics[width=8cm]{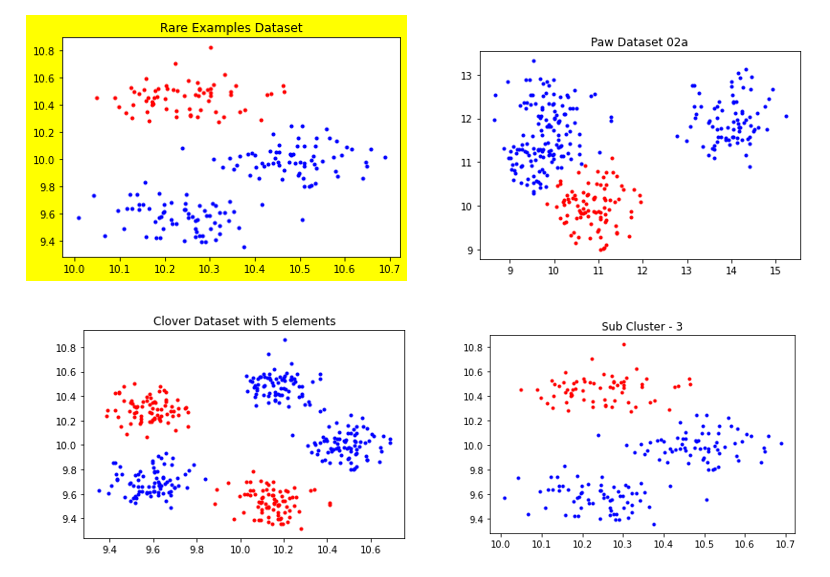}
    \caption{Datasets with subclusters}
    \label{fig:galaxy}
\end{figure}
\begin{figure}[htp]
    \centering
    \includegraphics[width=8cm]{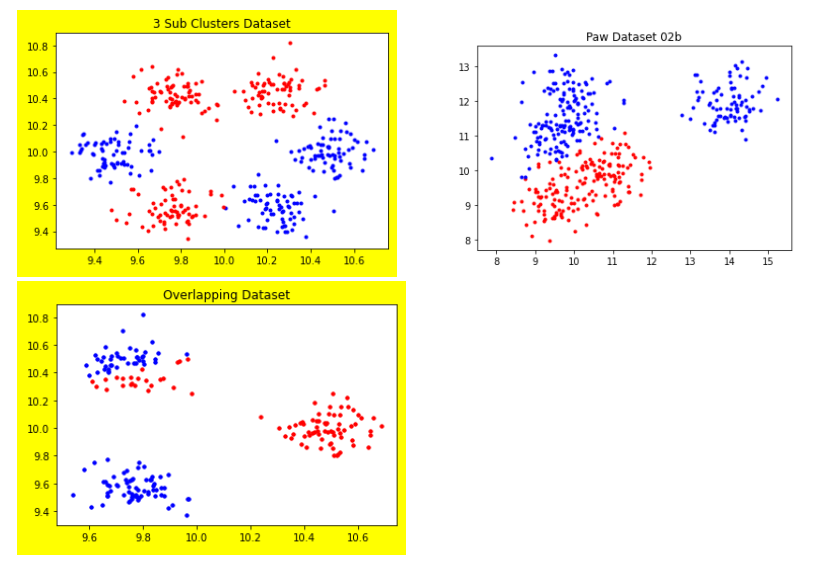}
    \caption{Overlapping dataset}
    \label{fig:galaxy}
\end{figure}
Below plots demonstrate the random data generation capability. Using the method we are able to generate data with multiple subclusters. For plotting the data we have using bandwidth estimation and MeanShift algorithm.
\subsection{Overlapping and Rare Examples}

Rare and overlapping examples are crucial parts of the dataset in real world situations. Most issues arise because of them and in order to make the method resilient to such situations we need to introduce such examples in the dataset. From the charts shown above you can see the algorithm’s capability to generate data overlap and introduce rare examples into the dataset.

\subsection{Improving the Performance of Classifiers}
We have chosen 4 methods which are simple random over-sampling, cluster based oversampling and nearest cleaning rule.

\section{Performance}

\subsection{Original Data Characteristics}
\begin{itemize}
    \item Number of samples: 400
    \item Number of features: 2
    \item Majority class label: 0
    \item Number of majority class samples: 316
    \item Minority class label: 1
    \item Number of Minority class sample: 84
    \item Imbalance Ratio : 3.8
\end{itemize}

\subsection{Data Characteristics after random Over Sampling}
\begin{itemize}
    \item Number of samples: 632
\item Number of features: 2
\item Majority class label: 1
\item Number of majority class samples: 316
\item Minority class label: 0
\item Number of Minority class sample: 316
\item Imbalance Ratio : 1.0
\end{itemize}

\subsection{Data Characteristics after NCR}
\begin{itemize}
   \item Number of samples: 258
\item Number of features: 2
\item Majority class label: 0
\item Number of majority class samples: 174
\item Minority class label: 1
\item Number of Minority class sample: 84
\item Imbalance Ratio : 2.1
\end{itemize}

\subsection{Data Characteristics after Cluster Based Oversampling}
\begin{itemize}
   \item Number of samples: 632
\item Number of features: 2
\item Majority class label: 1
\item Number of majority class samples: 316
\item Minority class label: 0
\item Number of Minority class sample: 316
\item Imbalance Ratio : 1.0

\end{itemize}

\section{Result and discussion}
Firstly, the impact of disturbing the borders of subregions in the minority class
was evaluated. It was simulated by increasing the ratio of borderline examples from
the minority class subregions. This ratio (further called the disturbance ratio) was
changed from 0 to 70
rable to the width of the sub-regions (sub-parts in data shapes).
The constructed classifiers were combined with the following focused pre-
processing methods:
\begin{itemize}
    \item Standard random oversampling (abbreviated as RO)
\item Japkowicz’s cluster oversampling (CO)
\item Nearest cleaning rule NCR
\end{itemize}

Cluster oversampling was limited to the minority class, and for baseline results (Base), both classifiers were ran without any
pre-processing.

Oversampling RO and CO performed comparably on all data sets, and on non-disturbed data sets they often over-performed focused method: NCR . If the overlapping area is large enough (in comparison to the area of the minority sub-clusters), and at least 30\% of examples from the minority class are located in this area, then oversampling method (NCR) strongly outperform random and cluster over-sampled with respect to sensitivity and G-mean.

In the first series of experiments we used data sets of three shapes, 800 examples and the imbalance ratio of 1:7. We also employed rule- and tree-based classifiers combined with the same pre-processing methods. In each training set 30\% of the minority class examples were safe examples located inside sub-regions, 50\% were located in the borderline/overlapping area (we denote them with B), and the remaining 20\% constituted the C rare examples.

NCR in handling rare examples located inside the majority class (also accompanied
with borderline ones). Such result has been in a way expected, as both methods were
introduced to handle such situations. The experiments also demonstrated that even
random oversampling better than NCR in classifying
safe examples from the minority class.

 The results of experiments showed that although NCR often led to the highest increase of sensitivity, at the same time it significantly deteriorated specificity and overall accuracy.  Moreover, it did not deteriorate the recognition of the majority classes as much as NCR.

\section{Conclusion}
The issue of learning classifiers using skewed data has been discussed. It's one of the most difficult subjects in data mining and machine learning.
Furthermore, it is still "open" from a theoretical standpoint, yet it is critical in a variety of application domains. On the other hand, a review of related literature reveals that it has sparked increased academic interest in the recent decade, with various specialized methods already developed. Despite the fact that some of them have been tested in experiments, additional general research concerns like the class imbalance, data features, and the competency of several popular approaches are still needed.

Other critical factors that affect classification performance include decomposition of classes into smaller sub-parts with too few examples (so-called small disjuncts), overlapping between classes (existence of too many borderline examples in the minority class), and the presence of rare or noisy examples located farther away from the decision boundary (deeper inside the distribution of the opposite class)

The effect of important parameters in artificial data sets on the performance of the most common rule-, tree-, and K-NN classifiers was investigated. First and foremost, these findings corroborated previous findings on the importance of the minority class decomposition into smaller sub-concepts based on simpler contrived data. We also found that increasing the subdivision of the class into small sub-parts decreased sensitivity or AUC values for more nonlinear decision boundaries.

Learning is extremely challenging due to the combination of class decomposition and overlapping (in particular for sensitivity measure and tree classifiers). Our investigations revealed that the quantity of questionable examples had a significant impact on the degradation of classification ability. In terms of sensitivity and G-mean measurements, concentrated re-sampling methods (such as NCR) outperformed random and cluster oversampling when the overlapping area was large enough.

\addtolength{\textheight}{-12cm}   




\section*{ACKNOWLEDGMENT}

Authors would like to thank ,  Dr. Shivani Gupta for their constant encouragement towards the realization of this work.


\end{document}